\documentclass[conference]{IEEEtran}
\IEEEoverridecommandlockouts
\usepackage{cite}
\usepackage{amsmath,amssymb,amsfonts}
\usepackage{algorithmic}
\usepackage{graphicx}
\usepackage{textcomp}
\usepackage{xcolor}
\usepackage[font=small,labelfont=bf]{caption}
\usepackage{subcaption}
\usepackage{lipsum} 
\usepackage{xargs}
\usepackage[inline]{enumitem}
\usepackage[T1]{fontenc}
\usepackage{balance}

\usepackage[colorlinks=true,urlcolor=blue]{hyperref}
\hypersetup{
    linkcolor = {black},
    citecolor = {black}
}

\usepackage{todonotes}

\begin{document}

\title{A Search Strategy and Vessel Detection in Maritime Environment Using Fixed-Wing UAVs}

\author{\IEEEauthorblockN{Marijana Peti$^{1}$, Ana Milas$^{1}$,  Natko Kra\v sevac$^{2}$, Marko Križmančić$^{1}$, Ivan Lon\v car$^{2}$,  \\  Nikola Mi\v skovi\'c$^{2}$, Stjepan Bogdan$^{1}$} \IEEEauthorblockA{\textit{$^{1}$Laboratory for Robotics and Intelligent Control Systems (LARICS),} \\ \textit{$^{2}$Laboratory for Underwater Systems and Technologies (LABUST),}\\ \textit{Faculty of Electrical Engineering and Computing, University of Zagreb,
Zagreb, Croatia}\\\textit{(email:
marijana.peti, ana.milas, natko.krasevac, marko.krizmancic, ivan.loncar, nikola.miskovic, stjepan.bogdan@fer.hr)}}}

\maketitle

\begin{abstract}
In this paper, we address the problem of autonomous search and vessel detection in an unknown GNSS-denied maritime environment with fixed-wing UAVs. The main challenge in such environments with limited localization, communication range, and the total number of UAVs and sensors is to implement an appropriate search strategy so that a target vessel can be detected as soon as possible.
Thus we present informed and non-informed methods used to search the environment. The informed method relies on an obtained probabilistic map, while the non-informed method navigates the UAVs along predefined paths computed with respect to the environment. The vessel detection method is trained on synthetic data collected in the simulator with data annotation tools.
Comparative experiments in simulation have shown that our combination of sensors, search methods and a vessel detection algorithm leads to a successful search for the target vessel in such challenging environments.
\end{abstract}

\begin{IEEEkeywords}
search strategy, fixed-wing UAV, detection, marine environment, MBZIRC, simulation
\end{IEEEkeywords}




\section{Introduction}
\label{introduction}
In a maritime environment, autonomous vehicles can be used for security, environmental monitoring, emergency operations such as search and rescue, or inspection of vessels and cargoes. Nowadays, poaching, smuggling, piracy, and human trafficking are becoming more common problems in such environments. Therefore, the security of maritime areas is of great importance.
Motivated by these problems ASPIRE \cite{ASPIRE}, the program management of Abu Dhabi's Advanced Technology Research Council, has organized the Mohamed Bin Zayed International Robotics Challenge (MBZIRC) 2023 \cite{MBZIRC}. The challenge aims to apply ground-breaking, state-of-the-art technologies to tackle the aforementioned problems.

The main idea of the challenge is to deploy a heterogeneous robotic system, composed of Unmanned Aerial Vehicles (UAVs), Unmanned Surface Vehicles (USVs), and a robotic arm attached to a USV, to collaboratively execute a mission consisting of inspection and intervention tasks. The inspection task consists of searching a given area and looking for a target vessel using UAVs, while the intervention task's goal is to approach the target vessel with the USV and pick up the required objects on the vessel with a UAV equipped with a manipulator. To make the whole competition even more challenging, several restrictions were introduced, such as the impossibility to use the Global Navigation Satellite System (GNSS) for positioning and navigation. Another restriction is the availability of only short-range communication among robots \cite{MBZIRC}. Some limitations on the total number of UAVs and sensors were also given, to make the simulation feasible.

The first phase of the competition consisted of submitting a white paper. In the second phase, the task is to complete the mission described above in the given simulator.  Our team, University of Zagreb, Faculty of Electrical Engineering and Computing (UNIZG-FER), is selected as one of the semi-finalists to participate in the second phase. Our team has been also selected as a top 5 finalist team for the third phase, where the implementation will be tested in a real-world environment.
This paper describes our team's second phase approach and solution to the inspection task, while the intervention task is described in \cite{Fausto} and \cite{Zoric}. 

\begin{figure}[t!]
 \centerline{\includegraphics[trim = {3cm 3cm 3cm 4cm}, clip, scale = 0.4]{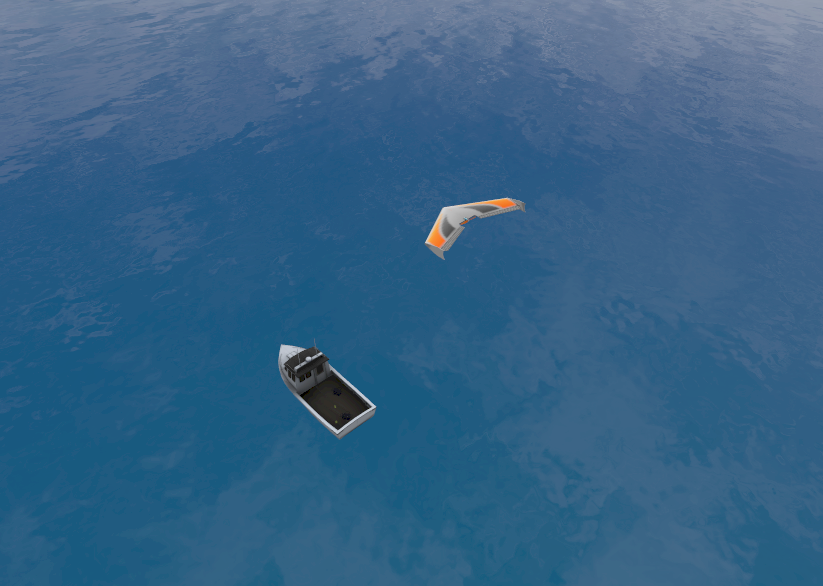}}
 \caption{A fixed-wing UAV flies in the search area above the vessel.}
 \label{fig:fw}
\end{figure}

As mentioned, the inspection task includes the search and detection of the target vessel with fixed-wing or quadrotor UAVs. The search zone is defined as 4 $\mathrm {km}^2$ area over the sea. Due to the limited mission time (60 minutes) and a relatively large area to be covered, fixed-wing UAVs are preferred over quadrotor UAVs. The fixed-wing UAVs are faster and have long range and endurance \cite{Yeong2015ARO}, therefore they are used for detection, coverage, and search task. The provided model of fixed-wing UAV \cite{FW_book} is shown in Fig. \ref{fig:fw}. 

There is plenty of literature related to UAV search in maritime environments, usually considered in search and rescue problems using multiple UAVs \cite{Sun2022}. In \cite{Cho2021}, the authors focus on solving the path planning problem for deploying multiple UAVs in a maritime area proposing a two-phase framework. OTOTE et al. \cite{AgbissohOTOTE2019} proposed a decision-making-oriented model to implement a maritime search and rescue plan with the support of optimal search theory.
The main difference in applying the same search algorithms to fixed-wing UAVs is their non-holonomic nature. When dealing with fixed-wing UAVs, the approaches should consider the motion constraints of such vehicles to plan a feasible trajectory. Recently, fixed-wing UAVs showed good performance when used for search and coverage tasks \cite{Sujit2014}.

Search can be performed with simple geometric flight patterns and more complex solutions based on full and partial information about the area \cite{drones_survey_2019}. Detailed comparison of different types of flight patterns is described in \cite{Andersen2014PathPF}. Instead of using predefined flight patterns, the authors in \cite {Wood2011} presented an approach to dynamically partition the target search area based on the target density distribution. 

Detecting objects in maritime environments is very versatile. In most cases, deep learning methods are used. In \cite{detection_rev}, authors describe an architecture of the system for UAVs patrolling above the sea, which could be easily integrated on an autonomous UAV. They tested it on different CNN architectures common in this field such as YOLOv3, Tiny YOLO, and faster R-CNN.

Regarding communication during a search in large environments, \cite{Sabo2014} proposes a heuristic for task allocation and routing for multiple UAVs with communication range being a key factor in the approach. Even though the authors included communication restrictions, there are no constraints on the GNSS signal, so our paper presents the design of the system for the search task in a maritime GNSS-denied environment. 


The contributions of this paper are summarized as follows:
\begin{enumerate}
 
    \item Control and path following algorithms for autonomous search in GSNN-denied maritime environments.
    \item The implementation and comparison of informed and non-informed methods used to search the environment.
    \item A vessel detection algorithm trained on synthetic data collection and integration into the search strategy.
    \item Comprehensive analysis and validation of the proposed approach in simulation.
\end{enumerate}

 The paper is organized as follows: Section \ref{sys-setup} describes the environment and each part of the system such as the UAV control, a localization algorithm, and a communication structure. Section \ref{search_strategies} presents two types of search strategies and explains how they are adapted to our system, while in Section \ref{detection}, a method for target vessel detection is introduced. In Section \ref{results} results are shown and the proposed approach is analyzed.  The paper ends with a conclusion in Section \ref{conclusion}.

\section{System Setup}
\label{sys-setup}

\subsection{Fixed-Wing}
\label{FixedWing}

\subsubsection{Control}
\label{Control}
\begin{figure}[t!]
 \centerline{\includegraphics[trim = {0.2cm 0.2cm 0.2cm 0.2cm}, clip, scale = 0.3]{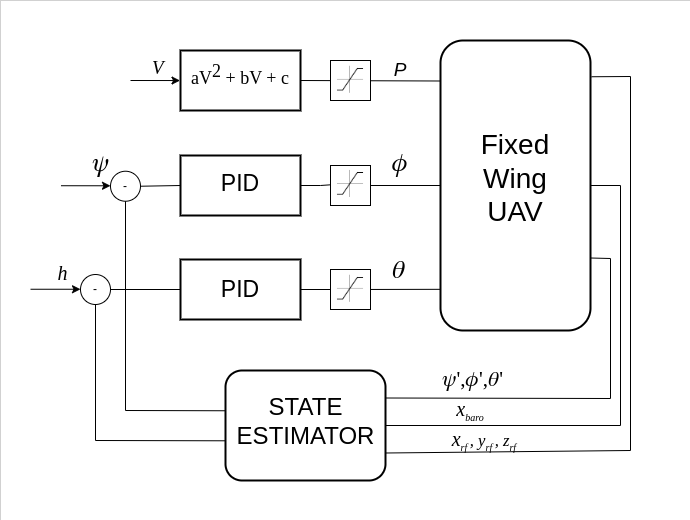}}
 \caption{Control scheme of a fixed-wing UAV.}
 \label{fig:fw-ctrl}
\end{figure}

For the given model of a fixed-wing UAV, the control inputs for the model in the simulator are roll ($\Phi$), pitch ($\theta$), and angular velocity of the propellers ($P$). Two uncoupled PID controllers are implemented to control the lateral and longitudinal motion of the UAV. The longitudinal motion is controlled by changing the pitch angle while the lateral motion (as well as yaw ($\Psi$)) is controlled by the roll angle.The angular velocity of the propellers  is used to set the velocity of the UAV ($V$). The relationship between the angular velocity of the propellers and the velocity of the UAV is determined experimentally by fitting the curve over a set of points. For the given angular velocity of the propellers, a simulation of the UAV is performed, while the ground velocity is calculated from the ground truth location data. The complete control scheme can be seen in Fig. \ref{fig:fw-ctrl}. 
Please note that for simplicity, the presence of wind is neglected and the angle of sideslip and angle of attack are assumed to be zero. Therefore, several simplifications can be introduced \cite{FW_book}:
\begin{enumerate*}
    \item The airspeed is equal to the ground speed,
    \item The course angle is equal to the heading angle.
\end{enumerate*}

\subsubsection{Feedback}
\label{Feedback}
The fixed-wing UAVs are equipped with an Inertial Measurement Unit (IMU) and a barometer. IMU provides measurements of attitude,  linear accelerations from the accelerometer, and rotation rates from the gyroscope. Given the precise measurements from the barometer and accelerometer in IMU, as well as the attitude angles from IMU ($\Phi$', $\theta$', $\Psi$'), the data is processed with a low-pass filter before being used to estimate the horizontal position $(x,y)$ in the linear Kalman filter (LKF) model. The barometer measurements ($X_{baro}$) are used to estimate the altitude. The prediction step of the LKF uses a constant acceleration model similar to \cite{Robi}:

\begin{gather}
\boldsymbol{\chi}_{k+1} = \mathrm{F}_k \boldsymbol{\chi}_k + \boldsymbol{\omega}_k, 
\label{eq:model_update} \\
\mathrm{F}_k = 
\begin{bmatrix}
\mathrm{C}_k & \boldsymbol{0}_{3\times3} \\
\boldsymbol{0}_{3\times3} & \mathrm{C}_k 
\end{bmatrix},
\label{eq:Fk} \\
\mathrm{C}_k = 
\begin{bmatrix}
 1 & \Delta T & 0.5\Delta T^2 \\
 0 & 1 & \Delta T \\
  0 & 0 & 1 \\
\end{bmatrix},
\label{eq:Ak}
\end{gather}
where $\Delta T$ is time difference between two steps ($k+1$ and $k$) and $\boldsymbol{\omega} \in \mathbb{R}^6$ represents process noise for the model states:

\begin{equation}
\boldsymbol{\chi}_{k} = \begin{bmatrix}
 x & \dot{x} & \ddot{x} & y & \dot{y} & \ddot{y}
\end{bmatrix} ^{T}.
\label{eq:model states}
\end{equation}

In addition to the standard sensors mentioned earlier, there is also a Radio Frequency (RF) range sensor and an HD camera on the front of the UAV. The RF sensor obtains range measurements used in the relative localization algorithm (see Section \ref{localization}), while the HD cameras are used for target detection (see Section \ref{detection}). Due to the lack of a velocity sensor, the only available measurement is position ($x_{rf}, y_{rf}$), which is then used to derive the velocity. Therefore, the observation model can be defined as follows: 

\begin{gather}
\boldsymbol{z}_k = \mathrm{H}_k \boldsymbol{\chi}_k + \boldsymbol{\upsilon}_k,
\label{eq:obs_update} \\
\mathrm{H}_k = 
\begin{bmatrix}
\mathrm{B} & \boldsymbol{0}_{3\times3} \\
\boldsymbol{0}_{3\times3} & \mathrm{B} 
\end{bmatrix},
\label{eq:Hk} \\
\mathrm{B} = 
\begin{bmatrix}
 1 & 0 & 0 \\
 0 & 1 & 0 \\
 0 & 0 & 1 
\end{bmatrix},
\label{eq:B}
\end{gather}
where $\boldsymbol{\upsilon} \in \mathbb{R}^6$ is measurement noise. Both $\boldsymbol{\omega}$ and $\boldsymbol{\upsilon}$ are considered zero-mean Gaussian distribution vectors with respective correlation matrices $\mathrm{Q} \in \mathbb{R}^{6\times6}$ and $\mathrm{R}\in \mathbb{R}^{6\times6}$. Diagonal components of $\mathrm{Q}$ are set to minimize estimated position and velocity noise and diagonal components of $\mathrm{R}$ are obtained from the relative localization algorithm. For the sake of brevity, equations of correction step in the Kalman filter are omitted.

\subsubsection{Path Following}
\label{PathFollower}
\begin{figure}[t]
 \centerline{\includegraphics[scale = 0.15]{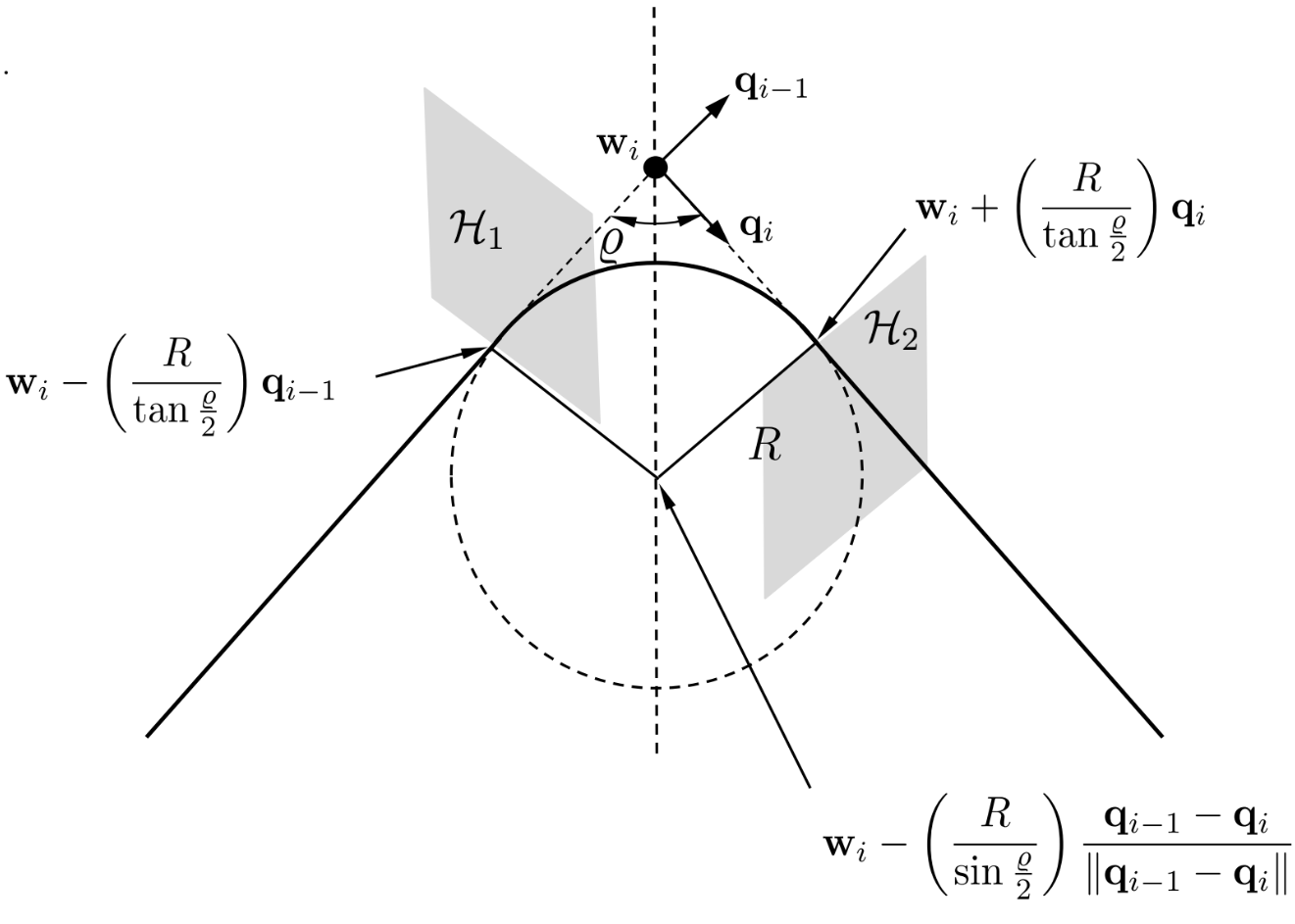}}
 \caption{Geometry of fillet path\cite{FW_book}. $w_i$ represents the current waypoint towards which UAV is heading, $R$ is the radius of the turn, $q_{i-1}$ is the unit vector from the previous waypoint to the current one ($w_i$) and it is also normal vector to the half-plane $H_1$. $q_i$ is the unit from the current waypoint to the one after $w_{i}$ and the normal vector of the half-plane $H_2$. In this way, when a UAV passes the half-plane $H_1$ it starts to turn to half-plane $H_2$ and heads toward the next point.}
 \label{fig:fillet}
\end{figure}

To ensure that the UAV follows the given waypoints, we utilize the method with fillet transitions (Fig. \ref{fig:fillet}) from \cite{FW_book}. In this way, instead of a usual sharp transition in the waypoint, the transition between two path segments is smoothed by defining the orbit and two half-planes as described in Fig. \ref{fig:fillet}.

In order to follow straight-line paths the desired yaw(heading angle) is calculated as follows: 
\begin{equation}
    \Psi = \mathrm{arctan2}(\Delta y, \Delta x) - s(2 / \pi) \mathrm{arctan}(e/k), 
\end{equation}

where $\Delta y$, and $\Delta x$ present defference in coordinates between next and previous waypoint, $s$ is path following parameter which presents approaching angle for following straight line and $e$ is the error from desired path orthogonal to the line connecting current two waypoints. $k$ is a gain for following straight-line path.

\subsection{Communication}
\label{communication}
\begin{figure}
    \centering
    \includegraphics[width=0.95\columnwidth]{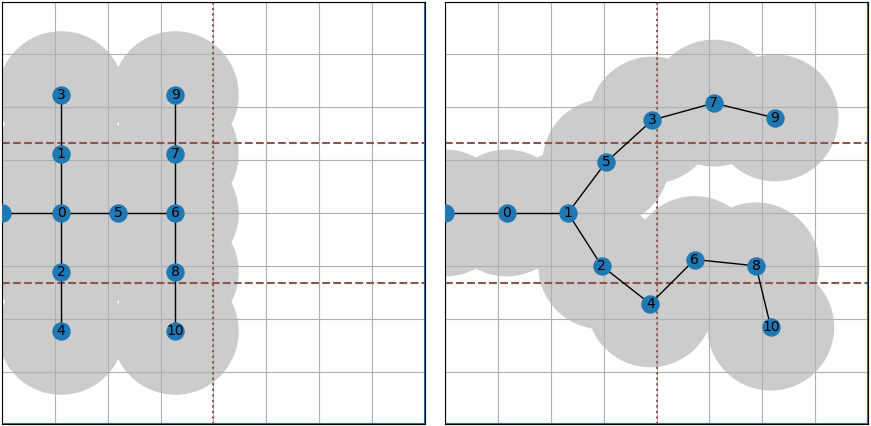}
    \caption{Positions, communication network, and coverage of relay UAVs for the first (left) and the second (right) half of the search procedure.}
    \label{fig:rel_configs}
\end{figure}
The communication plugin provided in the simulation enables two robots to exchange messages if they are within communication range $d_{max}=500$ m. Messages are serialized and sent over a single hop. The communication model imposes two limitations: 

\begin{enumerate}

    \item Messages are dropped with some probability $P_{drop}$ depending on their size in bytes $N_b$ and the distance between the robots $i$ and $j$ --- $d_{ij}$.  

    \item There is a maximum data rate allowed among robots communicating over the same network segment (e.g.:~1 Gbps). 

\end{enumerate} 

Transmitting power in the communication model is fixed and set to $Tx=25$ dBm. Received signal strength $Rx~\sim~\mathcal{N}(\mu,\,\sigma^{2})$ is a random variable drawn from a normal distribution with 
 
\begin{align} 
   &\mu = \left\{ 
       \begin{array}{ll} 
       -\infty, & d_{ij} > d_{max} \\ 
       Tx - l_0 + 10 \cdot f_e \cdot \log_{10}(d_{ij}), & d_{ij} \leq d_{max}, 
       \end{array}\right. \\ 
   &\sigma = 10, 
\end{align} 
where $l_0=40$ dBm is a constant describing the received power at $d_{ij}=1$ m, and $f_e=2.5$ is the fading exponent. 
The bit error ratio (BER) is calculated as  
\begin{equation} 
   BER = \mathrm{erfc}\left(\sqrt{10^{(Rx - Nf) / 10}}\right), 
\end{equation} 
where $\mathrm{erfc}(\cdot)$ is the complementary error function, and $Nf~=~-90$ dBm is constant noise floor\footnote{Hypotethical signal created from the sum of all the noise sources and unwanted signals within the system.}. 
 
Finally, the probability of a message being dropped is 
\begin{equation} 
   P_{drop} = 1 - e^{N_B \cdot \log(1 - BER)}. 
\end{equation}

\begin{figure*}[h]
     \centering
     \begin{subfigure}[b]{0.32\textwidth}
         \centering
         \includegraphics[width=\textwidth]{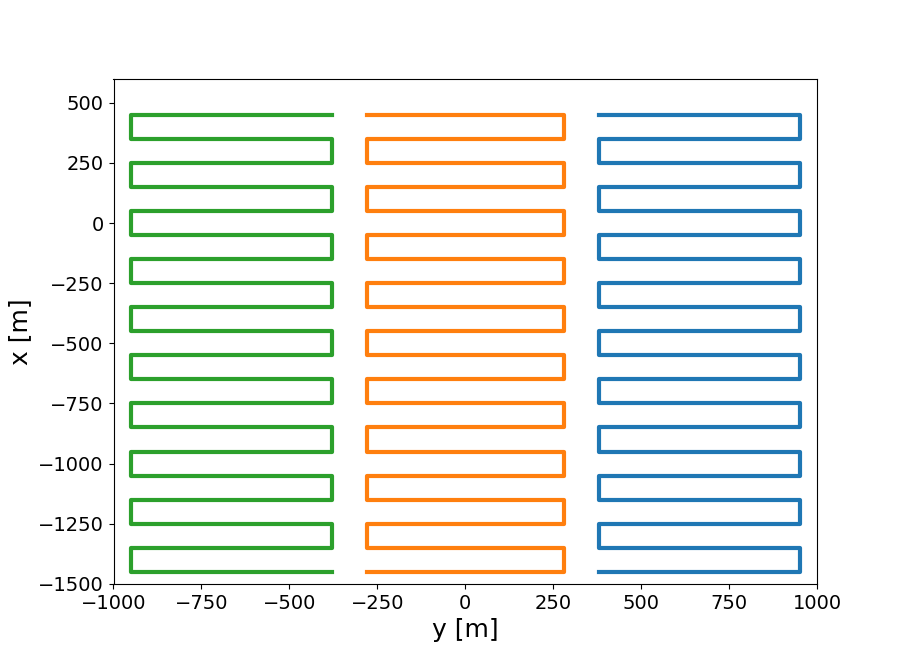}
         \caption{Parallel lines}
         \label{fig:Parallel}
     \end{subfigure}
     \begin{subfigure}[b]{0.32\textwidth}
         \centering
         \includegraphics[width=\textwidth]{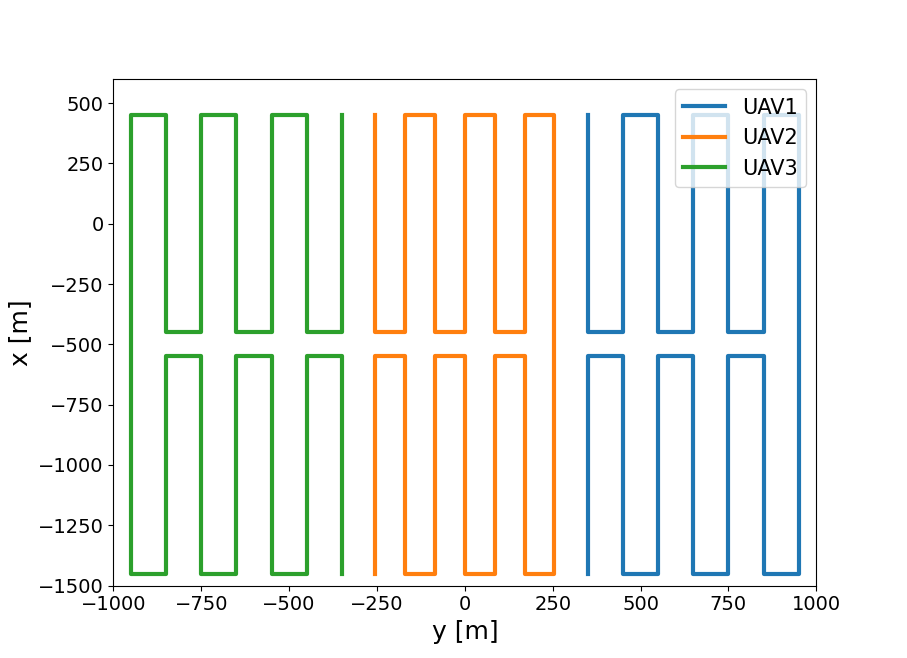}
         \caption{Creeping lines}
         \label{fig:Creeping}
     \end{subfigure}
     \begin{subfigure}[b]{0.32\textwidth}
         \centering
         \includegraphics[width=\textwidth]{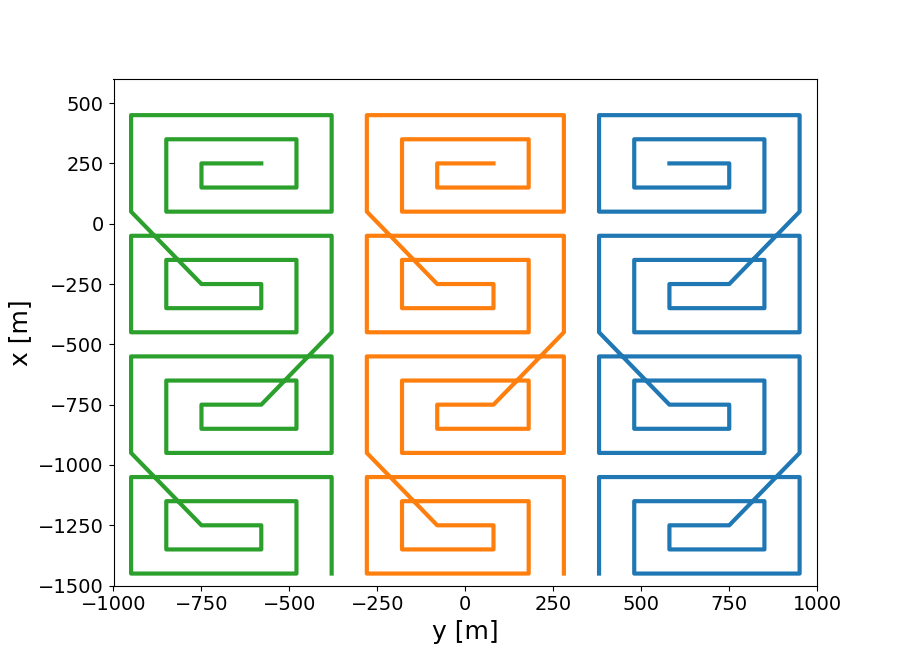}
         \caption{Spiral}
         \label{fig:Spiral}
     \end{subfigure}
        \caption{Proposed search patterns.}
        \label{fig:Patterns}
\end{figure*}

Given the communication model and its parameters, the effective communication range is severely limited. To ensure mostly uninterrupted communication between the fixed-wing UAVs and the base station during the entire search procedure, we set up a multi-hop network consisting of message-relaying quadrotor UAVs hovering at fixed locations throughout the search zone area. Following the limitations on the number of vehicles and sensors, there are not enough UAVs to cover the entire area. Therefore, we divided it into halves and defined two distinct relaying configurations shown in Fig. \ref{fig:rel_configs}. Once the fixed-wing UAVs almost complete the search in the first half, they send a signal to the relays to reposition themselves for the second part of the search pattern. To balance coverage and reliable communication, we set the distance between two relay UAVs to 300 m.

\subsection{Localization}
\label{localization}

Achieving good positioning of the system during mission execution requires absolute localization capabilities.
State estimation described in Section \ref{Feedback} relies on IMU and barometer.
The barometer is used to estimate absolute height assuming constant air pressure and calibration before UAV takeoff.
The IMU sensor is used for dead-reckoning $x-y$ estimation using accelerometer data. Due to the nature of the dead-reckoning procedure essentially relying on double integration of accelerometer data, the $x-y$ estimate drifts over time and the error is unbounded.

The RF sensor enables range estimation between two RF sensors mounted on different vehicles.
Range measurements between all agents are estimated simultaneously at a rate of 1~Hz. 
The maximum range is $\sim 1200$~m with $\sim1\%$ ranging error.

In the simulation, we utilized four static UAV vehicles (anchors later in the text) positioned at the starting gate to act as reference stations ("Tier 1") with known (initial) positions.
This allows vehicles in the RF range to estimate the distance to known locations in space.
Positioning anchors at different locations can be exploited to estimate their own position using established techniques (e.g., trilateration) if sufficient range measurements are available.
Spatial placement of anchors has a strong effect on localization performance with wider placement generally resulting in better precision.
In the simulation, we are restricted by the size of the runway area which is 50~m~$\times$~50~m, so we chose to place anchors in the corners of the runway.

During the mission execution, agents may go out of the range of the anchors, which would result in a loss of absolute localization.
To extend the localization range, a group of UAV can hover at predefined locations within the anchor range, which can act as new reference anchors ("Tier 2") with increased position uncertainty.
Vehicles within the range of "Tier 2" can also localize with even greater uncertainty to act as "Tier 3" anchors.
To mitigate this hierarchical localization structure, the location of the entire system can be determined simultaneously using relative localization algorithms by solving the equation described in the following section.

\subsubsection{Relative Localization Algorithm}

Relative localization relies on the exchange of measured distances between vehicles achieved by using the multi-hop communication scheme described in Section \ref{communication}. When information is aggregated each agent may find the solution locally.

The measurement model is given as follows:
\begin{align}
    d_{ij} = ||\textbf{a}_i - \textbf{a}_j|| + \epsilon_{ij}
\end{align}
where $\textbf{a}_i$ is $i$-th agent positions, $d_{ij}$ is distance measurement between agent $i$ and $j$, and $\epsilon_{ij}$ is normally distributed error in measured distance $d_{ij}$.

The system position state can be estimated by minimizing the following error:
\begin{align}
    \underset{a}{\text{minimize}} \sum_{i<j\leq A} \lambda_{ij} (d_{ij} - ||\textbf{a}_i - \textbf{a}_j||)^2,
\end{align}
where $A$ is the number of agents and $\lambda_{ij}$ is the weight with which distance error $\epsilon_{ij}$ affects the minimization function.

This problem is a basic metric multidimensional scaling problem \cite{groenen_multidimensional_2016}  well researched in literature. We chose to solve it by majorization to determine positions of all vehicles. 
The solution minimizes relative distances between agents which does not guarantee correct global placement.
Global placement can then be inferred by exploiting the last known state of the system's position state. The solution is then fed into the filter described in Section \ref{Feedback}. 
The uniqueness of the solution is also not guaranteed due to the sparsity of the distance matrix $D = [d_{ij}]$ due to local minima.
Note that the only challenging situation is when the matrix is sparse, which implies that the agents are not in the RF range of the anchors.


\section{Search Strategies}
\label{search_strategies}

In order to find the target vessel, non-informed and informed methods for searching the area are tested. The first one is based on covering the whole search area divided into equal parts for all UAVs and the second search method is based on an obtained probabilistic map. Due to the fact that UAVs are located in a GNSS-denied area, the whole area is split in half as explained in Section \ref{communication}. Therefore, these restrictions are taken into account when strategies are created.

\subsection{Non-informed method}
\label{non-inf}
In non-informed methods of area coverage, the main idea is to cover the whole search area in a specific geometric pattern. The most common patterns are back-and-forth and spiral \cite{drones_survey_2019}. Back-and-forth can be put into  parallel and creeping lines, while spiral can be round or square. The spiral pattern usually has the longest path and therefore takes the longest time, but it can easily be used for uniform area coverage \cite{drones_survey_2019}. In general, these patterns require low computational time to find search paths and can be easyly performed by UAVs. Three types of patterns are defined and tested in the simulation area: parallel lines, creeping spirals, and square spirals. The exact waypoints for each pattern can be found in Fig. \ref{fig:Patterns}.  
The maximum possible number of fixed-wing UAVs (due to limitations mentioned in Section \ref{introduction}) is 3. Therefore, in addition to the mentioned division, the whole area is divided into the same number of tracks.

\subsection{Informed method}
\label{inf}
In informed methods, the search strategy is usually based on a probabilistic map, based either on already known information about environmental conditions (sea currents, weather, to name a few)\cite{sea_curr_weather} or on a sensor such as radar. In this simulation, radar measurements are used. A USV with radar is docked to a base station. Simulated 3D radar has a range of 3.5 km and could cover the whole search area. It provides us with a probabilistic map of all vessels in that area. This probabilistic map is used to define paths for UAVs.

The order in which pionts are visited is determined by a $MinMax$ algorithm in a multiple traveling salesman problem. In this algorithm, the idea is to minimize the maximum length of the path among all UAVs. It is calculated separately for each half of the area to ensure that the fixed-wing and relay UAVs move to the second half at approximately the same time and repositioning is completed in the shortest possible time. Since the target vessel might move, when the UAV arrives in its vicinity, it performs a square spiral pattern with an increasing radius according to the field of view of the camera.

\begin{figure}[t!]
 \centerline{\includegraphics[scale = 0.4]{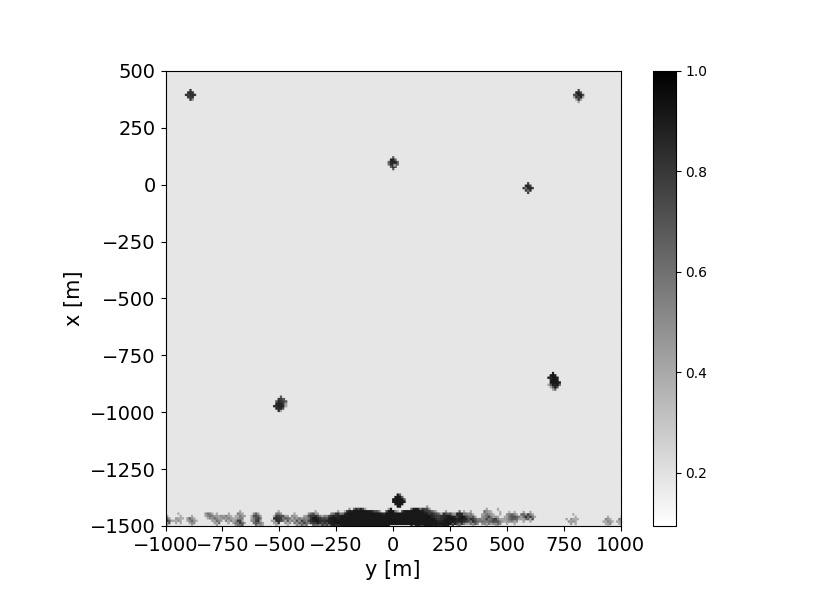}}
 \caption{An example of a probabilistic map of search area received after 30 s of scanning from radar. Darker spots represent a bigger probability of vessel location, while lower big black spot represents the coast.}
 \label{fig:probMap}
\end{figure}

\begin{figure}[t!]
 \centerline{\includegraphics[scale = 0.35]{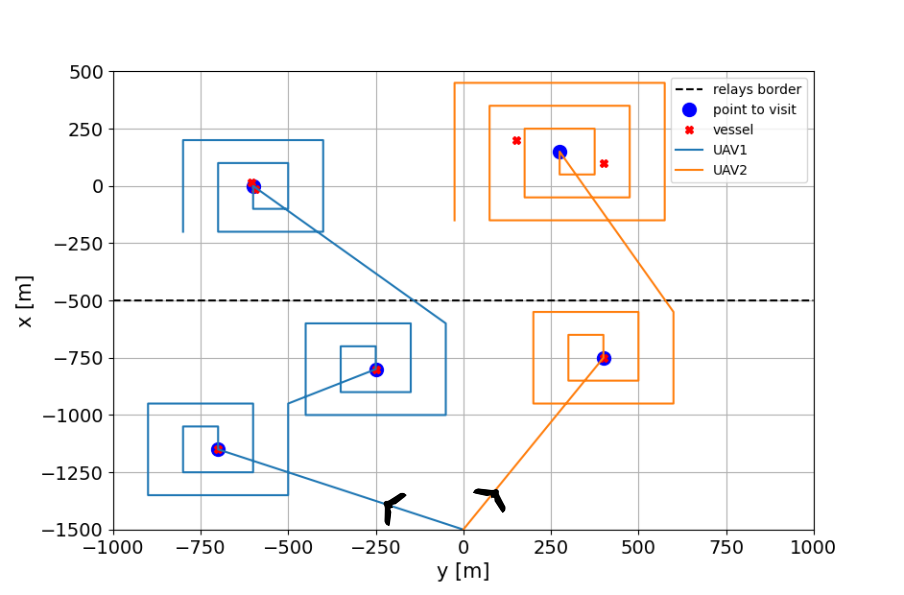}}
 \caption{An example of waypoints for fixed-wing UAVs. The black dashed line represents the boundary between two halves, red marks are the positions of vessels detected by radar and blue dots represent points to visit. In most cases, the vessels are located in the same place as the blue dots, but if vessels are very close to each other (upper left and right spirals), the spirals are enlarged to cover all close-positioned vessels.}
 \label{fig:wp}
\end{figure}

\section{Vessel Detection}
\label{detection}
\begin{figure}
 \begin{tabular}{cc}
  \includegraphics[width=40mm]{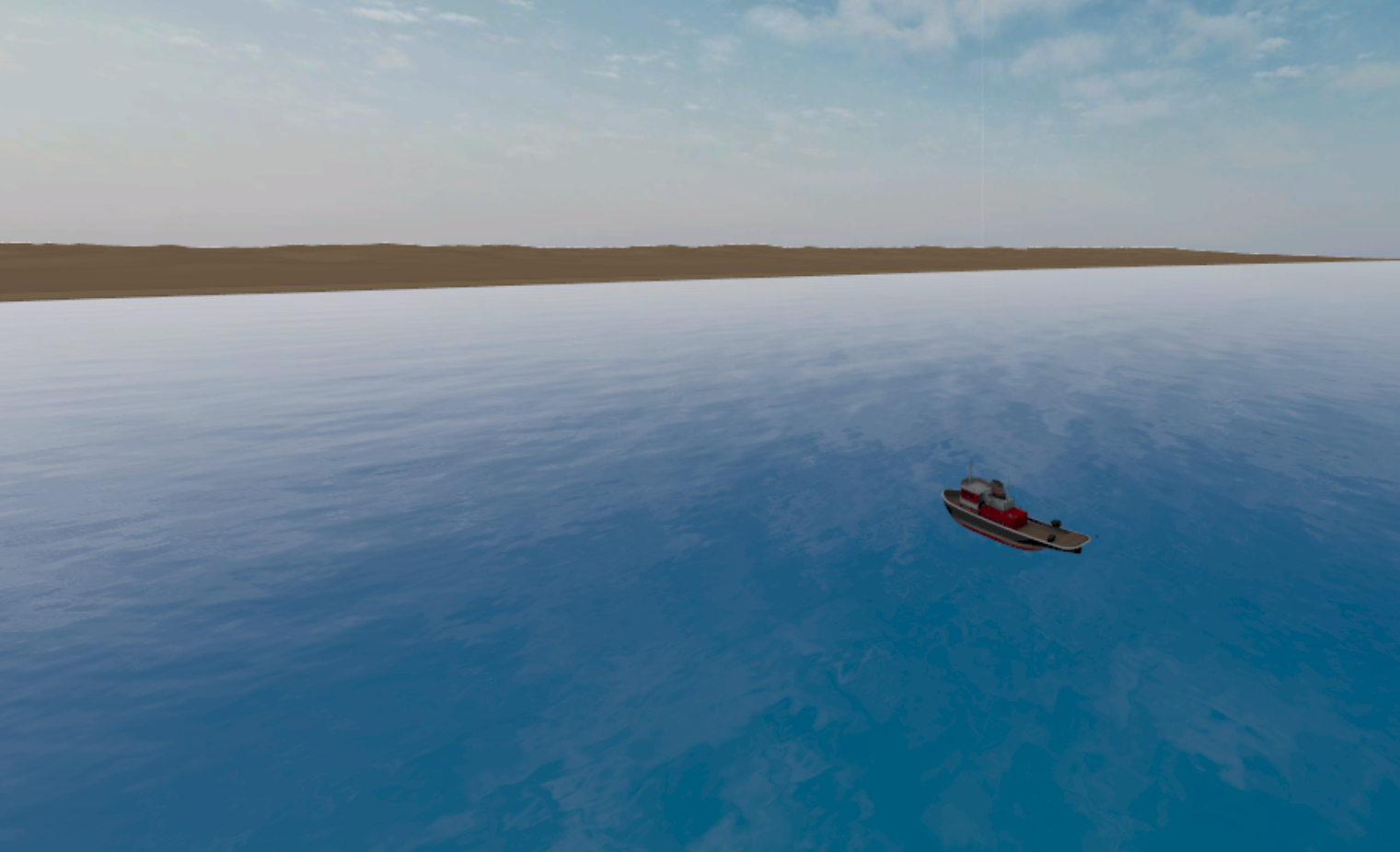} &   \includegraphics[width=40mm]{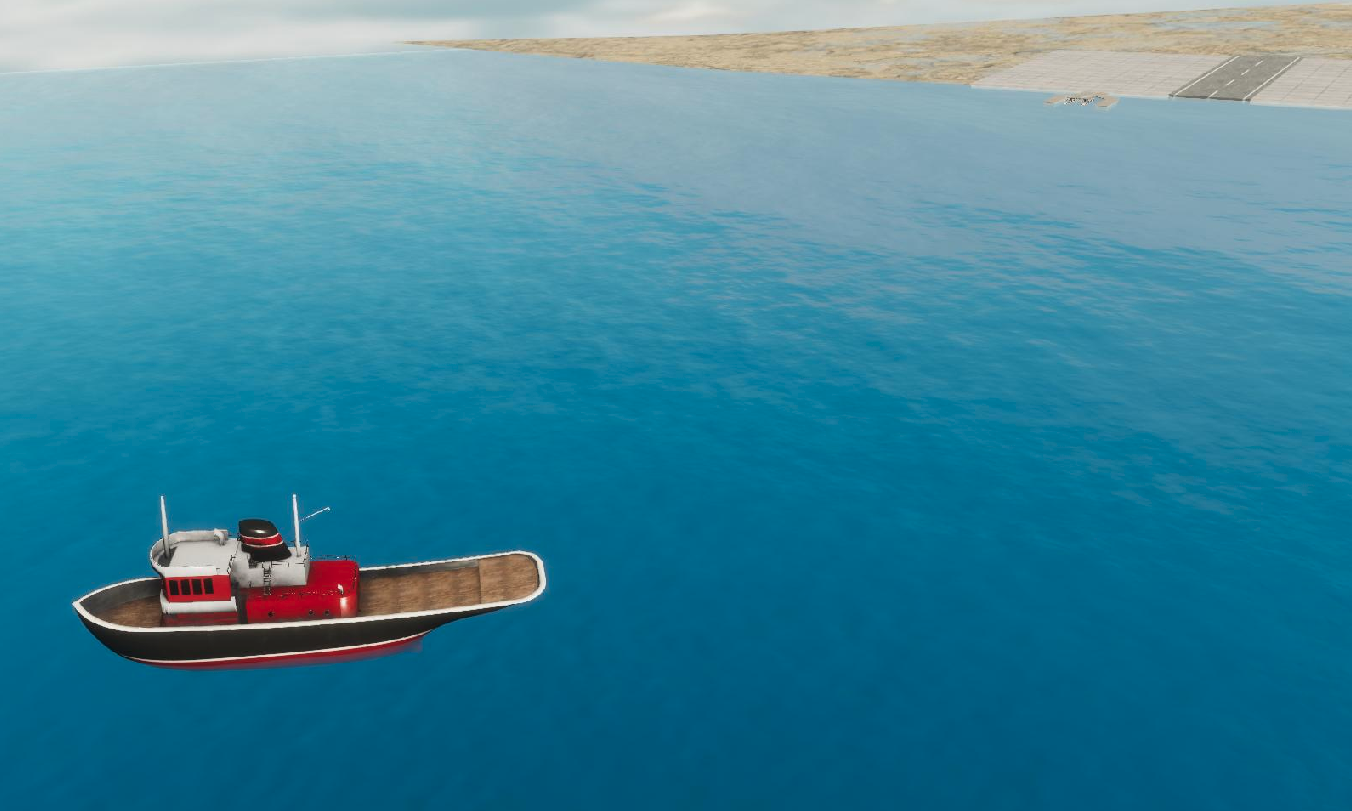} \\
\end{tabular}
 
 \caption{Comparison between MBZIRC scene (left) and MARUS scene (right)}
\label{fig:sim_comparison}
\end{figure}

In order to detect a target vessel, the fixed-wing UAV flying over the search area is equipped with an HD camera. It is the long-range HD camera, mounted on the front of the UAV and tilted downward at a 20$^{\circ}$ angle.
Current methods for object detection tasks from the camera image, together with vessel detection, are mostly based on deep learning \cite{patel2022deep, kim2018probabilistic}.  This requires a large amount of previously annotated images of vessels in different lighting conditions and from different angles. While there are some publicly available data sets for ship detection \cite{8438999,zheng2020mcships} with provided experimental results of popular
baseline detectors, none of those are applicable to our detection task because of the difference in vessel classes and major visual gap between simulation and real-world images. In addition to that, with no data collection tools in the MBZIRC simulator, we make use of the MARUS simulator \cite{9976969} and its tools for fast and easy collection of synthetic data sets. In order to use the aforementioned tools, the scene from the MBZIRC simulator had to be recreated in MARUS. When training a deep neural network, it is very important that the training data reflects the real data to which the network will be applied. Since the training data is acquired in the MARUS simulator and applied to the MBZIRC simulator, a problem called domain shift \cite{SunFS15} can occur. The difference in image statistics between the training data and the data the network is used on can cause the network to perform well on the validation data and poorly in the final application. We minimize this difference by recreating a new scene similar to the original scene in the MBZRIC simulator. This involves matching the color and reflectivity of the sea, as well as the color of the sky and the density of the clouds. In addition, since the camera is mounted on the fixed-wing UAV, the training data should be captured from the air with the same field of view and rotation of the camera. By recreating the scene very similarly, as seen in Fig. \ref{fig:sim_comparison}, domain shift will be easier to overcome. 

There are seven possible vessel types, one of which is selected as the target vessel for the task. 3D models of all vessels are imported into the scene and a training dataset is collected. In total, about 10 000 images are then divided into subsets for training, validation and testing.

For training, YOLOv5 \cite{yolov5} is used because it provides simple and fast learning with good performance in terms of accuracy. We selected the weights trained on COCO data set\cite{cocodataset} since one of the classes in this dataset is a ship. The backbone weights are frozen, meaning that already trained weights are used to detect a ship, and other weights are trained on the previously collected dataset.


\section{Results}
\label{results}
In order to choose the best and the fastest search strategy, first, the time to cover the whole area with three search patterns is compared, mentioned in \ref{non-inf}. In all the experiments UAVs have the velocity of 25 m/s and flies at 100 m height and the exact waypoints are determined based on the camera field of view (at this height it is $100 \text{ m} \times 250 \text{ m}$. The times are acquired after running the simulation for each pattern three times and the average time is calculated (Table \ref{tab:patterns_percentage}).

It can be seen that the times needed to cover the area do not differ a lot. The time of flying the spiral pattern is a little bit longer, which is expected since this type of pattern has a longer path \cite{drones_survey_2019}. Another criterion in choosing the search pattern is the availability of the signal during the search and the percentage of messages received from multiple relays, which improves total localization. For this use case, it was decided to compare the percentage of total messages received from more than five relays. Results can be seen in Table \ref{tab:patterns_percentage}.
\begin{table}[ht]
\centering
\caption{Comparison of the time required for fixed-wing UAVs to cover the whole area depends on the search pattern, percentage of the time when UAV is in communication range and percentage of the time UAV receives messages from more than 5 relays while doing the covering of the whole area.}
\begin{tabular}{|c| c c c|} 
 \hline
 Pattern & Parallel & Creeping & Spiral \\ 
 \hline
 Simulation time [min] & 10.1 & 10.0 & 11.2 \\ 
 \hline
In-the-range time [\%] & 89.1  & 85.9 & 88.4 \\ 
 \hline
 $>$ 5 relay [\%] & 54.3  & 49.7  & 39.7 \\ 
 \hline
\end{tabular}
\label{tab:patterns_percentage}
\end{table}
For parallel and spiral search patterns, the percentage of time UAVs receive data is similar, but only the parallel pattern has the highest percentage of the time that  UAVs receive data from more than 5 relays. The main reason for this is that this is the only search pattern from the mentioned ones here that does not have a straight line at the edge of the search area, where UAVs receive messages from fewer relays. 

\begin{figure}
 \centerline{\includegraphics[scale = 0.15]{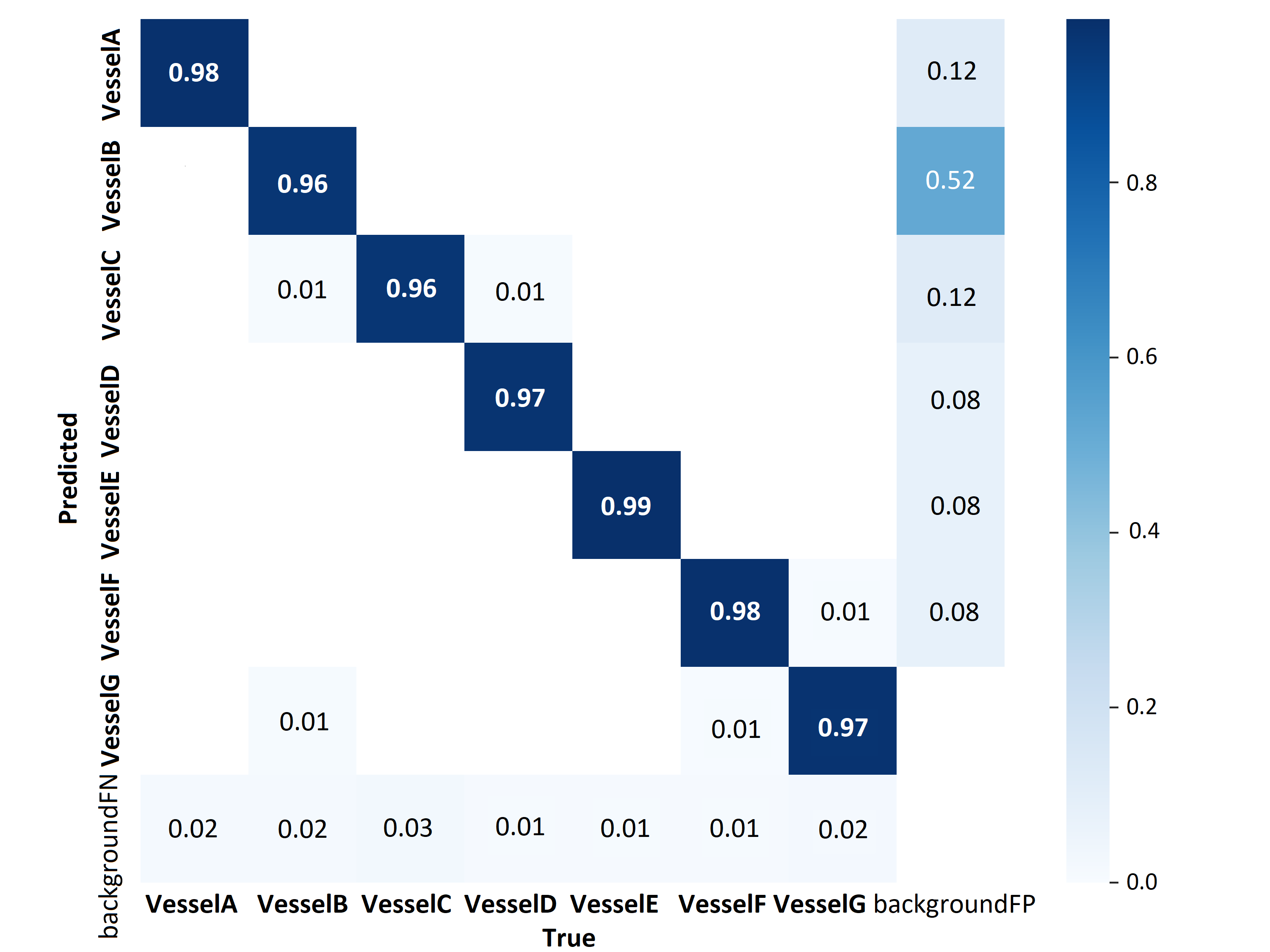}}
 \caption{Confusion matrix of the final model run on test subset.}
 \label{fig:confusion_matrix}
\end{figure}

\begin{figure}
\begin{tabular}{cc}
  \includegraphics[width=40mm]{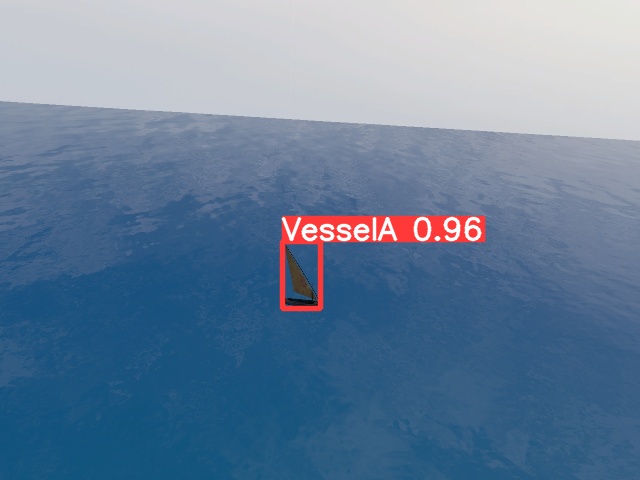} &   \includegraphics[width=40mm]{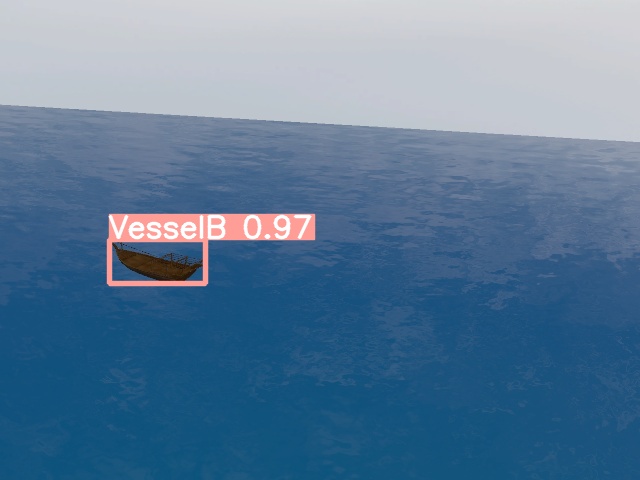} \\
(a) Vessel A & (b) Vessel B \\[3pt]
 \includegraphics[width=40mm]{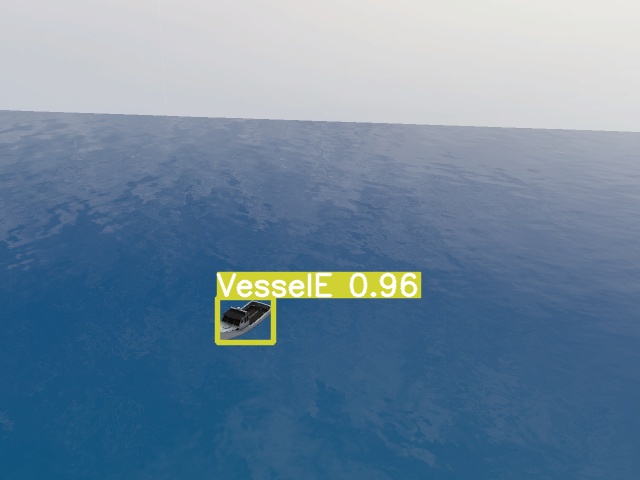} &   \includegraphics[width=40mm]{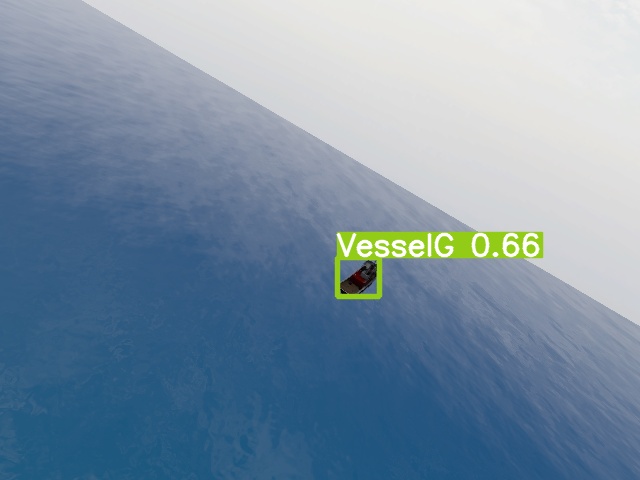} \\
(c) Vessel E & (d) Vessel G \\[3pt]
\end{tabular}
\caption{Vessel detection visualized with bounding box and model confidence.}
\label{fig:detection}
\end{figure}

In our case, the target vessel is set to be vessel E. Thus, more images for the target vessel (vessel E) are collected so it dominated the dataset in a number of instances. 
 The training is performed for 150 epochs and the final results are presented in form of a confusion matrix in Fig. \ref{fig:confusion_matrix}. This matrix shows that the model performs best for the target vessel E, as only 1\% is not detected. Results also show that the model has minor problems with vessel C as it is detected in a very small percentage as either vessel B or vessel D. In summary, the network successfully learned to identify and detect each of the seven possible vessel types in the test subset. Running the network in the MBZIRC simulator shows that domain adaptation is successful as shown in Fig. \ref{fig:detection}. 

\begin{figure}[t]
 \centerline{\includegraphics[scale = 0.6]{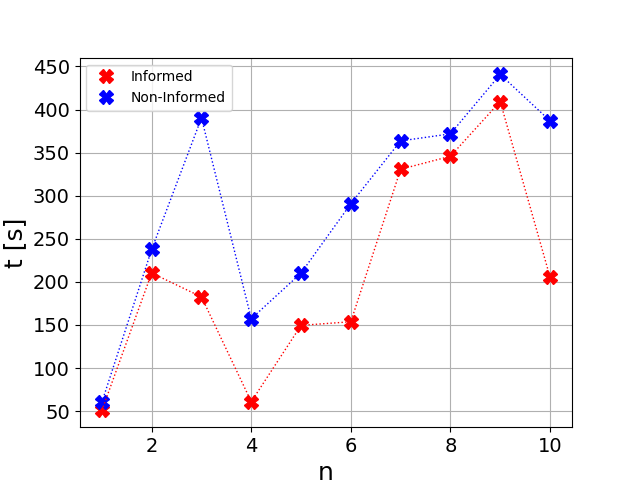}}
 \caption{Time comparison of detection of a target vessel in the informed and non-informed method over $n=10$ runs.}
 \label{fig:comparison}
\end{figure}

If we look back to Table \ref{tab:patterns_percentage}, when executing the parallel search pattern, only turns are performed at the edges, so the UAVs do not spend much time in the areas with reduced communication capabilities. Therefore, for our system, this parallel search pattern is chosen in the non-informed method. This technique is then compared to the method described in Section \ref{inf}, where radar is used to obtain information about the environment. 
Due to the possibility of exceeding the total number of payload sensors allowed, the final number of fixed-wing UAVs had to be reduced when the radar is added.  Therefore, the informed method is tested with 2 fixed-wing UAVs, and the time required to find the target vessel is compared with the non-informed method with a parallel search pattern and 3 fixed-wing UAVs. Due to the demanding system in the simulation, a total of 10 experiments are performed with randomly generated positions of vessels in the search area. The time required to detect the target vessel is compared and can be seen in Fig. \ref{fig:comparison}.
The time to find the target vessel is reduced when using the informed method, by using radar and only 2 fixed-wing UAVs. By adding the radar, a smaller amount of area needs to be covered by UAVs and therefore it takes less to find the target vessel.

\section{Conclusion and Future Work}
\label{conclusion}
In this work, we described our approach to tackle the demanding task of the MBZIRC competition in simulation. The first part of this maritime challenge was to find a moving target vessel in a GNSS-denied area with restricted communication. Fixed-wing UAVs are used for scanning the area due to their higher flight speed, while multirotor UAVs facilitated long-range communication and localization. In order to detect the vessel, we  created a synthetic dataset of different vessel types in various lighting conditions and viewing angles and used it to train a YOLOv5 detector. The whole system was tested in simulation with two approaches for scanning the competition area. Both informed and non-informed methods showed promising results. However, the informed method had a faster time in finding the target due to the use of a probabilistic map obtained from the radar.  
Future work will address the problem of control under windy conditions and its application in real-world experiments.

\section*{Acknowledgements}
\small{This work has been supported in part by the European Union through the European Regional Development Fund - The Competitiveness and Cohesion Operational Programme (KK.01.1.1.04.0041) through Horizon Europe CSA project AeroSTREAM - Strengthening Research and Innovation Excellence in Autonomous Aerial Systems, grant agreement No. 101071270. Research work presented in this article has been supported by the project "Razvoj autonomnog besposadnog višenamjenskog broda" project (KK.01.2.1.02.0342) co-financed by the European Union from the European Regional Development Fund within the Operational Program "Competitiveness and Cohesion 2014-2020". The content of the publication is the sole responsibility of the project partner UNIZG-FER.
The work of doctoral students Ana Milas and Marko Križmančić has been supported in part by the “Young researchers’ career development project--training of doctoral students” of the Croatian Science Foundation funded by the European Union from the European Social Fund. }

\bibliographystyle{ieeetr}
\typeout{}
\balance
\bibliography{UT23}
\end{document}